\documentclass[journal]{IEEEtran}

\IEEEoverridecommandlockouts                              

\usepackage{graphicx}
\usepackage{color}
\usepackage{bm}
\usepackage{amsmath}
\usepackage{amsfonts}
\usepackage{subcaption}
\usepackage{caption}
\usepackage{url}
\usepackage{xcolor}
\usepackage{makecell}
\usepackage{algorithm}
\usepackage{algpseudocode}
\usepackage{float}
\newcommand{\rev}[1]{\textcolor{black}{#1}}

\title{\LARGE \bf Towards Robot Skill Learning and Adaptation with Gaussian Processes
}

\author{A K M Nadimul Haque, Fouad Sukkar, Sheila Sujipto, Cedric Le Gentil, \\Marc G. Carmichael and Teresa Vidal-Calleja 
\thanks{Corresponding author e-mail: akmnadimul.haque@student.uts.edu.au}
}
\begin{document}

\maketitle
\thispagestyle{empty}
\pagestyle{empty}

\begin{abstract}

General robot skill adaptation requires expressive representations robust to varying task configurations. 
While recent learning-based skill adaptation methods refined via Reinforcement Learning (RL), have shown success, existing skill models often lack sufficient representational capacity for anything beyond minor environmental changes. In contrast, Gaussian Process (GP)-based skill modelling provides an expressive representation with useful analytical properties; however, adaptation of GP-based skills remains underexplored. This paper proposes a novel, robust skill adaptation framework that utilises GPs with sparse via-points for compact and expressive modelling. The model considers the trajectory's poses and leverages its first and second analytical derivatives to preserve the skill's kinematic profile. We present three adaptation methods to cater for the variability between initial and observed configurations. Firstly, an optimisation agent that adjusts the path's via-points while preserving the demonstration velocity. Second, a behaviour cloning agent trained to replicate output trajectories from the optimisation agent. Lastly, an RL agent that has learnt to modify via-points whilst maintaining the kinematic profile and enabling online capabilities. Evaluated across three tasks (drawer opening, cube-pushing and bar manipulation) in both simulation and hardware, our proposed methods outperform every benchmark in success rates. Furthermore, the results demonstrate that the GP-based representation enables all three methods to attain high cosine similarity and low velocity magnitude errors, indicating strong preservation of the kinematic profile. Overall, our formulation provides a compact representation capable of adapting to large deviations from a single demonstrated skill.

\end{abstract}

\section{INTRODUCTION}

Robots performing manipulation tasks in industrial settings frequently encounter variations in task configurations (TCs), such as changes in object pose or start and final end-effector poses. Effective task execution under such variability requires not only modelling the skill but also adapting it to new TCs. Skill modelling typically provides a compact representation that captures the underlying motion. Skill adaptation updates the skill model to satisfy new TCs. However, it is often desired to not only adapt the path but also retain the kinematic profile of the demonstrated skill. In this work, we define this type of adaptation as structured skill adaptation. While recent work has improved robustness to task variability~\cite{haque2024constrained}, structured skill adaptation in highly varying TCs remains a challenge.

\label{sec:intro}
\begin{figure}[ht]
    \centering
    \includegraphics[width=.9\linewidth, height=0.85\linewidth]{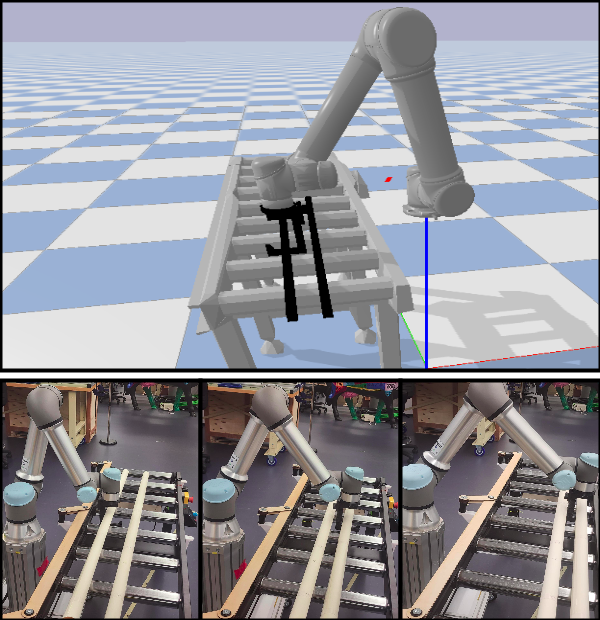}
    \caption{The agent learns to adapt the skill from a single demonstration (top) for one-shot deployment on a real robot under various initial conditions (bottom).}
    \label{teaser}
\end{figure}

End-to-end models have been proposed for skill learning and adaptation; however, their reliance on large datasets and training times limits real-world applicability \cite{chen2025trajectory}. In contrast, learning from demonstration (LfD) frameworks efficiently learn skills from expert demonstrations through dynamical systems to parametrise trajectories~\cite{ravichandar2017learning}, enabling skill adaptation to varying TCs~\cite{haque2024constrained, nematollahi2022robot}. 
Hybrid approaches combining skill models with learning are gaining traction for their robustness to varying TCs and sensor noise. State-of-the-art works combine parameterised models like Dynamic Movement Primitives (DMPs), Probabilistic Movement Primitives (ProMPs) and Gaussian Mixture Models (GMMs) with deep reinforcement learning (DRL). However, DMPs cannot capture trajectory distributions, GMMs often fail to capture complex motions, and ProMPs require high-dimensional weight vectors to represent complex motion, limiting compactness and scalability. In contrast, \rev{Gaussian Processes (GP)} offer stronger representational power~\cite{arduengo2023gaussian}, analytical derivatives \cite{sarkka2011linear} and compact parameterisation \cite{sukkar2023robotic}, making them well suited for structured skill adaptation. 
While GPs have been applied to skill learning~\cite{jaquier2020learning, sukkar2023robotic}, their potential for structured skill adaptation remains unexplored. 

Most GP-based skill parameterisation methods use dense formulations that adapt pose trajectories via new observations \cite{arduengo2023gaussian}, often resulting in distorted kinematic profiles. While nonlinear constraint optimisation can preserve the kinematic profile, as demonstrated later in this work, its slow convergence limits real-time deployment. Moreover, unconstrained adaptation of GP models can lead to deviations in the kinematic profile. \rev{Consequently, preserving the structure of a demonstrated skill remains an open challenge, and existing shape preservation methods still fall short for real-world deployment~\cite{akbulut2021acnmp, an2024robust}.} 

\begin{figure*}
    \centering
    \includegraphics[width=1\linewidth]{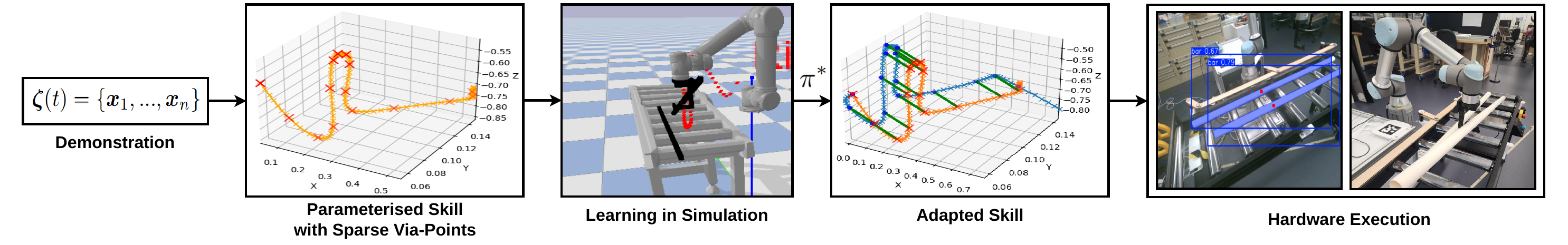}
    \caption{A single demonstration is parameterised through a GP formulation with sparse via-points. The subsequent skill adaptation method learns to tune the via-points to accomplish the task in varying TCs. This can be successfully transferred to hardware for one-shot execution.}
    \label{pipeline}
\end{figure*}

This paper proposes a novel GP-based robust, structured skill adaptation framework using sparse via-points for efficient skill modelling from a single demonstration. Fig.~\ref{pipeline} provides an overview of the framework, in which a single demonstration is parameterised using GPs defined by a sparse set of via-points. Online adaptation is achieved by conditioning the GP on newly observed TCs through the strategic placement of key via-points, such as start or final poses. A major advantage of this GP-based parameterisation is its differentiability, enabling analytical derivatives of the pose trajectory to be queried and used to preserve the underlying kinematics. This representation further enables fine-tuned adaptation policies to be trained in simulation and transferred to real hardware in one shot. \rev{We evaluate the proposed framework on three manipulation tasks in both simulation
and on a real robot, one of which is illustrated in Fig.~\ref{teaser}.} An algorithmic overview is given in Algorithm~\ref{algo:pipeline_general}.

We formulate three adaptation methods: (i) an optimisation-based method that adjusts via-points to match both the demonstrated trajectory and its kinematic profile; (ii) a behaviour cloning model trained to imitate the optimisation method in real time; and (iii) an RL policy trained from scratch, guided by a signature similarity reward metric rather than explicit supervision from the optimisation-based method.

To summarise, the contributions are as follows: 
\begin{itemize}

    \item A novel structured skill adaptation framework that employs a GP-based parametrisation using sparse via-points.
    \item An optimisation-based method that retains the kinematic profile of the demonstration in varying TCs. 
    \item A behaviour cloning-based method that runs in real-time and successfully imitates the output of the expert optimisation-based method.
    \item An RL policy exploration method guided by a signature similarity metric to retain the kinematic profile without explicit supervision from the optimisation-based method.    
    \item Successful sim-to-real transfer demonstrating skill adaptation for real-world object manipulation.
\end{itemize}
\section{Related Work}
\label{sec:literature}
In skill learning, demonstrated trajectories are often parameterised to enable generalisation to new TCs. DMPs use attractor dynamics and forcing functions to reproduce motions~\cite{ijspeert2013dynamical}. Adapting DMPs to varying TCs often requires external policies~\cite{lu2021incremental, davchev2022residual}. However, DMPs prioritise smoothness over precision, which can hinder accurate skill execution \cite{luciani2024imitation}.

Many state-of-the-art works adopt probabilistic trajectory parameterisation methods such as GMMs. GMMs represent trajectories as a weighted combination of Gaussian distributions, which can be tuned to adapt to different TCs using an external framework \cite{nematollahi2022robot, haque2024constrained}. While lightweight, their limited modelling capacity and flexibility \cite{arduengo2023gaussian} limit successful skill adaptation to newly observed TCs, mostly near the predictive mean \cite{nematollahi2022robot, haque2024constrained}. Some works have suggested the use of GMMs in conjunction with DMPs \cite{dong2023robot} and GP Regression \cite{jaquier2020learning}. The former proposes skill parameterisation with GMMs and using it to initialise their DMP model to reproduce a smooth motion; however, skill adaptation was not addressed. The latter proposes using a GMM to initialise a GP framework, enabling the GP to follow Cartesian position predictions that are outside the training data \cite{jaquier2020learning}. ProMPs have emerged as a more powerful alternative to GMMs and DMPs, and have been combined with RL for adaptation \cite{carvalho2022adapting}. However, parameterisation with high-dimensional weights increases complexity in RL integration, leading the authors of \cite{carvalho2022adapting} to adapt only the nominal trajectory instead of the skill model itself. 

GPs are flexible, kernel-based, nonlinear probabilistic regression models that enable an expressive representation of a trajectory. Since GPs can be analytically differentiable depending on the kernel choice~\cite{sarkka2011linear}, the kinematics of the motion can also be analytically obtained, finding uses in accurate motion planning~\cite{mukadam2016gaussian, mukadam2018continuous}.
Since GPs are more proficient in regressing a single output, researchers often opt for separate GPs per axis \cite{arduengo2023gaussian}. Additionally, GP-parameterised trajectories can easily be adapted to new TCs with a via-point-based parameterisation~\cite{sukkar2023robotic}.
Despite their flexibility, introducing discrete via-points can reduce the GP model accuracy and change the demonstrated skill's shape and velocity profile. 

Outside the GP-based frameworks, alternative approaches have suggested using explicitly defined loss functions to retain key properties, such as the shape and geometry of the demonstration. Authors in \cite{akbulut2021acnmp} learn a low-dimensional latent space from demonstrations through a Conditional Neural Process, before using a combination of supervised and RL to adapt a skill. \rev{While this implicitly retains the spatial structure of the demonstrated trajectory, the skill's kinematic profile is not preserved.} An alternative approach leverages a Generative Adversarial Network equipped with a shape preservation loss to generate letter-shaped trajectories \cite{an2024robust}. However, this is computationally expensive and requires extensive supervision. Furthermore, both of these formulations operate in a latent space that is difficult to interpret and does not explicitly retain the skill kinematics. In contrast, GPs offer a principled and efficient alternative for kinematic preservation due to their smooth interpolation and analytical derivatives, yet they remain unexplored in the context of structured skill adaptation.

\section{Background}
\label{sec:background}
\subsection{Trajectory Parameterisation with Gaussian Process}
Let us consider a demonstrated trajectory defined as a set of robot poses over time $\bm{\zeta} (t)$ modelled using a GP as:
\begin{equation}
    \bm{\zeta}(t) \sim \mathcal{GP}(0, \bm{k}(t, t')),
    \label{eq:gpshort}
\end{equation}
where $\bm{k}(t,t')$ represents the kernel function that encodes the covariance between various time instances of the trajectory. Given $N$ noisy observations of trajectory poses $\bm{y}_i = \bm{\zeta}(t_i) + \bm{\eta}$, where $\eta \sim \mathcal{N}(0, \sigma^2_y\bm{I})$ and $i = 1, 2, ..., N$, the multi-variate Gaussian distribution for the noisy observations $\mathbf{y}$ and the latent function value at a new query point 
$\bm{\zeta}(\hat t)$ is given by, 
\begin{equation}
\begin{bmatrix}
\mathbf{y} \\
\bm{\zeta}(\hat t)
\end{bmatrix}
\sim \mathcal{N}\left(\begin{bmatrix}
0 \\
0
\end{bmatrix},
\begin{bmatrix}
\mathbf{K}_{tt} + \sigma_y^2 \mathbf{I} & \mathbf{k}_t(\hat{t})^\top \\
\mathbf{k}_t(\hat{t}) & \bm{k}(\hat{t},\hat{t})
\end{bmatrix}\right),
\label{eq:gplong}
\end{equation}
where $\mathbf{k}_t$ represents the cross-correlation between the observations and the query point and $\mathbf{K}_{tt}$ is the covariance matrix of the observations. Then, the posterior mean of the trajectory at the query point can be expressed as, 
\begin{eqnarray}
    \bm{\zeta}(\hat{t}) &= \mathbf{k}_t(\hat{t}) \left[\mathbf{K}_{tt} + \sigma_y^2 \mathbf{I}\right]^{-1} \mathbf{y}.
    \label{eq:gpmean}
\end{eqnarray}

Given a new observation $\bm{y}_o$ at time $t_o$, the GP can be updated by appending $(t_o, \bm{y}_o)$ to the existing training set $\mathbf{y}$. The posterior mean trajectory can then be recomputed using Eq.~\eqref{eq:gpmean}, ensuring the updated trajectory passes through the observed location. 

Linear operators are often used to extend GP models, enabling the system to analytically query the GP mean function to obtain properties of the underlying function, such as the derivative, integral, Jacobian, and Hessian.

In this work, we are interested in the first and second derivatives given by the following, 
\begin{equation}
    \dot{\bm{\zeta}}(t) = L_{\bm{\zeta}}^t \bm{k}_t(t) \left( \bm{K}_{tt} + \sigma_y^2 \bm{I} \right)^{-1} \mathbf{y}
    \label{eq:gp_first_deriv_operator}
\end{equation}
\begin{equation}
    \ddot{\bm{\zeta}}(t) = L_{\bm{\zeta}}^{tt} \bm{k}_t(t) \left( \bm{K}_{tt} + \sigma_y^2 \bm{I} \right)^{-1} \mathbf{y}
    \label{eq:gp_second_deriv_operator}
\end{equation}
where \( L_{\bm{\zeta}}^t = \frac{\partial}{\partial t} \) and \( L_{\bm{\zeta}}^{tt} = \frac{\partial^2}{\partial t^2}\) are the first and second derivative operators respectively.
Note that Eqs. \eqref{eq:gp_first_deriv_operator} and \eqref{eq:gp_second_deriv_operator} only require differentiating the kernel function. Thus, instead of numerically evaluating the derivatives of the trajectory to obtain velocity and acceleration, as commonly done, this work analytically infers these quantities by simply querying the linearly operated GP at the desired time $t$.

\subsection{Soft Actor Critic}
Soft Actor-Critic (SAC) \cite{haarnoja2018soft} is a stochastic, off-policy reinforcement learning algorithm designed for continuous action spaces. Unlike conventional RL methods, SAC promotes exploration by maximising entropy, defined as the expected negative log-probability of actions.
The policy $\pi$ is updated by maximising a reward $R$ (as a function of current state $s_t$ and next state $s_{t+1}$ and the action $a_t$) augmented by a discount factor $\gamma$ and the entropy term, given by,
\begin{equation}
       \pi^* = \arg \max_{\pi} \underset{\tau \sim \pi}{\mathbb{E}}\left[ \sum_{t=0}^{\infty} \gamma^t \left( R(s_t, a_t, s_{t+1}) + \alpha H\left(\pi(\cdot|s_t)\right) \right) \right],\label{eq:SAC_policy_update}
\end{equation}
where $\alpha$ is the temperature parameter controlling the exploration-exploitation trade-off, $H$ is the entropy,  and $\mathbb{E}$ the expectation taken over the policy and environment.

\section{Towards Skill Learning and Adaptation}
\label{sec:proposed}

\subsection{Skill Parameterisation with GP}

Given a demonstrated end-effector trajectory $\bm{\zeta}(t)$, we can model the motion with six independent GPs paired with their timestamps, i.e., 
\begin{equation}
\begin{aligned}
    \mathbf{p}(t) &\sim \mathcal{GP}(0, \bm{k}(t, t')) \\
    \mathbf{r}(t) &\sim \mathcal{GP}(0, \bm{k}(t, t')),
\end{aligned}    
\label{eq:inducings}
\end{equation}
where $\mathbf{p}(t) = \mathrm{[p_x, p_y, p_z]^T}$ is the end-effector position, and $\mathbf{r}(t) = \mathrm{[r_x, r_y, r_z]^T}$ is the end-effector orientation expressed with the rotation vector parameterisation. Following \cite{sukkar2023robotic}, we have opted for the Squared Exponential (SE) kernel, where its hyperparameters have been empirically chosen to best fit the provided demonstration. In practice, we found length-scales of $\lambda=0.66$ and noise of values $\sigma_y^2=0.005$ worked well for all targeted skills.

\rev{To facilitate skill adaptation without retraining, we follow the work in \cite{sukkar2023robotic} and introduce a set of sparse via-points for each of the six GPs, linearly spread over time. These sparse sets of via-points summarise our GP parametrised trajectory.}
Thus, we can rewrite Eq.~\eqref{eq:gpmean} to query points along the trajectory as,
\begin{equation}
    \begin{aligned}
    \bm{p}^{\ast}(t) &= &\mathbf{k}_{t}(t) \left[\mathbf{K}_{tt} + \sigma_y^2 \mathbf{I} \right]^{-1} \bm{\Gamma}_p \\
    \bm{r}^{\ast}(t) &= &\mathbf{k}_{t}(t) \left[\mathbf{K}_{tt} + \sigma_y^2 \mathbf{I} \right]^{-1} \bm{\Gamma}_r,
    \end{aligned}
\label{eq:new_inducing}
\end{equation}
where $\bm{\Gamma}_p$ and $\bm{\Gamma}_r$ are the matrices of via-points for position and orientation, respectively. Furthermore, by applying linear operators on the GP kernel, the corresponding linear velocities and rotation-vector derivatives ($\dot{\bm{p}}^{\ast}(t), \dot{\bm{r}}^{\ast}(t)$) and second derivatives ($\ddot{\bm{p}}^{\ast}(t), \ddot{\bm{r}}^{\ast}(t)$) are obtained through Eq.~\eqref{eq:gp_first_deriv_operator} and Eq.~\eqref{eq:gp_second_deriv_operator} respectively.

To ensure that the sparse points represent the demonstrated trajectory, we formulate a non-linear optimisation that minimises the differences between the true posterior and the posterior approximated with the via-points through Eq.~\eqref{eq:new_inducing} as,
\begin{equation}
\begin{aligned}
\{\bm{\tilde{\Gamma}_{p_x}}, \bm{\tilde{\Gamma}_{p_y}}, \bm{\tilde{\Gamma}_{p_z}}\} &= \underset{\{\bm{\tilde{\Gamma}_{p_x}}, \bm{\tilde{\Gamma}_{p_y}}, \bm{\tilde{\Gamma}_{p_z}}\}}{\arg\min} \left( \sum_{i=1}^{Q} \|\mathbf{p}_i^{\ast}(t_i) - \mathbf{p}_i\|^2 \right) \\
\{\bm{\tilde{\Gamma}_{r_x}}, \bm{\tilde{\Gamma}_{r_y}}, \bm{\tilde{\Gamma}_{r_z}}\} &= \underset{\{\bm{\tilde{\Gamma}_{r_x}}, \bm{\tilde{\Gamma}_{r_y}}, \bm{\tilde{\Gamma}_{r_z}}\}}{\arg\min} \left( \sum_{i=1}^{Q} \|\mathbf{r}_i^{\ast}(t_i) - \mathbf{r}_i\|^2 \right),
\label{eq:new_inducing_detailed}
\end{aligned}
\end{equation}
where $\mathbf{\tilde\Gamma}$ are the optimised via-points, and \rev{$Q$} is the number of via-points selected to enforce the approximation. This ensures that the inference using $\bm{\tilde{\Gamma}_{p}}$ and $\bm{\tilde{\Gamma}_{r}}$ accurately produces the demonstrated skill.

\subsection{Skill-GP: Nonlinear Optimisation for Skill Adaptation}
The via-point-based parameterisation enables efficient adaptation to new TCs, such as start, object and goal location, by conditioning the GP on the new observations. These locations can be incorporated into the GP by aligning these locations with a subset of via-points from the original set of via-points, ensuring the adapted skill passes through the observed locations. This subset depends on the structure of the demonstration. 

Although simple skills can be adapted this way, complex, realistic motions require fine adjustments of the rest of the via-points to preserve the shape and kinematics of the skill. Thus, we propose an optimisation formulation to minimise the motion discrepancies. Specifically, we formulate a non-linear least squares problem similar to Eq. \eqref{eq:new_inducing_detailed}, such that it minimises discrepancies in the first and second derivatives of position and rotation-vector trajectories of the adapted skill with respect to the demonstration. Furthermore, to prevent the optimised points from reverting to their original values, we leverage the newly observed via-points by anchoring the optimisation around those points. 
We designate a small subset of via-points, $\bm{\Gamma^{\text{fixed}}}$, to reflect observed conditions, typically two or three points out of 15 in each axis, such as the start location, contact point and final location. These via-points can be identified by calculating the nearest via-point from the observed conditions.
The remaining via-points, $\bm{\Gamma^{\text{free}}}$, are then optimised to satisfy the desired trajectory. Thus, the equation becomes, 
\begin{equation}
\begin{aligned}
\{\bm{\Gamma_{p_x}^{\ast}}, \bm{\Gamma_{p_y}^{\ast}}, \bm{\Gamma_{p_z}^{\ast}}\} 
&= \underset{\{\bm{\Gamma_{p_x}^{\text{free}}}, \bm{\Gamma_{p_y}^{\text{free}}}, \bm{\Gamma_{p_z}^{\text{free}}}\}}{\arg\min} 
\sum_{i=1}^{Q} \left\| \dot{\mathbf{p}}_i^{\ast}(t_i) - \dot{\rev{\tilde{\mathbf{p}}}}_i \right\|^2 \\
&\quad + \lambda \sum_{i=1}^{Q} \left\| \ddot{\mathbf{p}}_i^{\ast}(t_i) - \ddot{\rev{\tilde{\mathbf{p}}}}_i \right\|^2 \\
\{\bm{\Gamma_{r_x}^{\ast}}, \bm{\Gamma_{r_y}^{\ast}}, \bm{\Gamma_{r_z}^{\ast}\}} 
&= \underset{\{\bm{\Gamma_{r_x}^{\text{free}}}, \bm{\Gamma_{r_y}^{\text{free}}}, \bm{\Gamma_{r_z}^{\text{free}}}\}}{\arg\min} 
\sum_{i=1}^{Q} \left\| \dot{\mathbf{r}}_i^{\ast}(t_i) - \dot{\rev{\tilde{\mathbf{r}}}}_i \right\|^2 \\
&\quad + \lambda \sum_{i=1}^{Q} \left\| \ddot{\mathbf{r}}_i^{\ast}(t_i) - \ddot{\rev{\tilde{\mathbf{r}}}}_i \right\|^2,
\end{aligned}
\label{eq:velocity_optimization_general}
\end{equation}
where, $\lambda$ is a scaling factor, determined empirically, $(\tilde{\mathbf{p}}_i, \tilde{\mathbf{r}}_i)$ are the demonstrated
end-effector pose samples from $\bm{\zeta}(t)$ at time $t_i$.
With these optimised via-points, we can infer the adapted skill $\bm{\zeta}'(t)$ through Eq.~\eqref{eq:new_inducing}.

\begin{algorithm}[h]
\caption{Structured Skill Adaptation (General Pipeline)}
\label{algo:pipeline_general}
\begin{algorithmic}[1]
\Require Demonstration $\bm{\zeta}(t)$, task configuration $\mathcal{TC}$
\Ensure Adapted trajectory $\hat{\bm{\zeta}}(t)$
\State Parameterise $\bm{\zeta}(t)$ with a GP defined with sparse via-points
\State Extract task-relevant via-points from $\mathcal{TC}$

\State Condition the GP on the observed via-points

\State Adapt the via-points and Update the GP posterior

\State Query the updated GP to reconstruct $\hat{\bm{\zeta}}(t)$

\State \Return $\hat{\bm{\zeta}}(t)$
\end{algorithmic}
\label{algo:pipeline}
\end{algorithm}
\subsection{Skill Cloning: Skill Adaptation with Imitation Learning}
Although we can solve Eq.~\eqref{eq:velocity_optimization_general} through standard non-linear least squares methods, it is unsuitable for real-time applications due to their high computational cost and slow convergence to the global minimum. Thus, we propose to train a policy through behaviour cloning (BC) \cite{kumar2021should} that mimics the behaviour of the optimisation-based method. 

Let us formulate the problem as a supervised state-action learning problem, defined by the state space, $ \mathcal{X} $, and the action space, $ \mathcal{A}$. Each BC state $\bm{s}^{BC}  \in \mathcal{X}$, is defined as the concatenation of two distance vectors: the vector from the end-effector to the object, $\bm{d}_o^{BC}$, and vector from the object to its goal pose, $\bm{d}_g^{BC}$.
 
The corresponding actions, $\bm{a}^{BC} \in \mathcal{A}$, are the proposed shifts $\Delta \bm{\Gamma}$, to the via-points of our GP formulation,
\begin{equation}
    \bm{s}^{BC} = \{ \bm{d}_o^{BC}, \bm{d}_g^{BC} \}; \bm{a}^{BC} = \Delta \bm{\Gamma}.
    \label{eq:rl_states}
\end{equation}
Given an expert policy $\pi^*(\bm{s}^{BC})$, the goal is to learn a policy that imitates the actions of the expert given the state inputs. Essentially, the objective function minimises, 
\begin{equation}
\mathcal{L}_{BC} = \sum_{t=1}^{N} \left\| \pi^*(\bm{s}^{BC}) - \bm{a}^{BC} \right\|\bm{^}2.
\label{eq:BC_loss}
\end{equation}
Since Skill-GP can achieve excellent adaptation from a single expert demonstration, we can treat it as the pseudo-expert policy, and create a diverse dataset of the updates to the via-points given by Skill-GP at a given state observation.
Namely, we run Skill-GP for a certain number of episodes and collect the state-action pairs of those runs to build up our expert dataset. Using the expert dataset, we learn to imitate the optimisation in a supervised manner to obtain our cloned policy. We choose a shallow network of similar architecture to that of the SAC \cite{haarnoja2018soft}. This facilitates efficient, real-time skill adaptation in a diverse set of TCs.
\subsection{GPRL: Fast Off-Policy Skill Adaptation}
While imitation learning offers a practical means of policy learning by mimicking expert behaviour, it relies heavily on building a large and diverse expert dataset. This can be time-consuming, especially in varying environments, as Skill-GP requires significant time to converge and generate expert trajectories. As an alternative, we train an RL policy with a single demonstration to adaptively refine the parameterised skill while preserving its kinematic profile. 
\subsubsection{Problem Formulation}
Let us augment our previous formulation of the problem as a complete MDP by adding the transition probability distribution $ \mathcal{P} $, the reward function $ r: \mathcal{X} \times \mathcal{A} \rightarrow \mathbf{R} $, and the discount factor $\gamma \in [0, 1] $. \rev{At time-step $t_s$, the state is given by $ \bm{s}_{t_s}^{GPRL} \in \mathcal{X} $ as the concatenation of: the distance vector from the end-effector to the object to be manipulated $\bm{d}_o^{GPRL} $, the distance vector from the object to its goal pose $\bm{d}_g^{GPRL}$, the time-step $t_s$, the means of the current velocities and accelerations $\bar{\bm{v}}$ and $\bar{\bm{a}}$. Note that we use our formulation to analytically query velocity and acceleration via Eqs.~\eqref {eq:gp_first_deriv_operator} and \eqref {eq:gp_second_deriv_operator}, then compute their means to be included as state inputs, yielding a 19-dimensional state-space vector. We propose a global adaptation by updating the GP, thus the mean velocity and accelerations provide a compact representation of the current kinematic information.
The actions, $\bm{a}_{t_s}^{GPRL} \in \mathcal{A}$, at each time-step, $t_s$, are again the proposed shifts $\Delta \bm{\Gamma}$ to the via-points of our formulation, which are typically small to ensure stable axis-angle updates,}
\begin{equation}
    \bm{s}_{t_s}^{GPRL} = \{ \bm{d}_o^{GPRL}, \bm{d}_g^{GPRL}, t_s,\bar{\bm{v}},\bar{\bm{a}} \}; \bm{a}_{t_s}^{GPRL} = \Delta \bm{\Gamma}.
    \label{eq:rl_states}
\end{equation}
The aim is to find an optimal policy, $\pi^\ast$, that maximises the cumulative reward through Eq.~\eqref{eq:SAC_policy_update}.
To ensure the adapted GP trajectory, $\bm{\zeta}'(t)$, inferred through Eq.~\eqref{eq:new_inducing} using the updated via-points $ \bm{\Gamma^\ast} = {\bm{\tilde{\Gamma}}} + \Delta \bm{\Gamma}$, retains the shape and velocity profile of the skilled demonstration, $\bm{\zeta}(t)$, we must ensure that the new kinematic profile follows the provided demonstration as closely as possible. Through Eq.~\eqref{eq:gp_second_deriv_operator}, we can infer the velocity of the updated trajectory as $\dot{\bm{\zeta}}'(t)$ and the provided demonstration as, $\dot{\bm{\zeta}}(t)$. Thus, our goal becomes finding a policy $\pi^\ast(\bm{a}|\bm{s})$ such that $\mathrm{min} |\dot{\bm{\zeta}}'(t)-\dot{\bm{\zeta}}(t)|$ is satisfied.
\subsubsection{Shape Preservation with Signature Similarity Scores}
To guide the policy towards retaining the velocity profile of the demonstration, we must provide compact and consistent scalar rewards based on the velocity similarities. We propose the use of signature similarity scores \cite{barcelos2024path}, a metric that is derived from signature kernels \cite{salvi2020computing}. Given a trajectory $\bm{\zeta}(t)$, the path signature encodes its geometry and order through a collection of iterated integrals that capture increasingly complex relations between the dimensions of the path, making it a compact and rich representation of its dynamics and geometry. The signature kernel allows us to compare the signatures of two trajectories. In our case, given the velocities of the skilled demonstration and updated trajectory are defined as $\dot{\bm{\zeta}}(t)$ and $\dot{\bm{\zeta}}'(t)$ respectively, the signature kernel is formulated as,
\begin{align}
k_{\text{sig}}\left(\dot{\bm{\zeta}}(t), \dot{\bm{\zeta}}'(t)\right) = \left\langle S\left(\dot{\bm{\zeta}}(t)\right), S\left(\dot{\bm{\zeta}}'(t)\right) \right\rangle_{\mathcal{H}}
\label{eq:signature_kernel}
\end{align}
where $S(\cdot)$ denotes the truncated path signatures, and $ \mathcal{H}$ is the corresponding Hilbert space, capturing iterated integrals of the path that encode its sequential structure and dynamics.
To obtain a bounded similarity score from 0 to 1, the normalised signature index is calculated as,
\begin{equation}
    \text{Sim}\left(\dot{\bm{\zeta}}(t), \dot{\bm{\zeta}}'(t)\right) = \frac{k_{\text{sig}}\left(\dot{\bm{\zeta}}(t), \dot{\bm{\zeta}}'(t)\right)}{\sqrt{k_{\text{sig}}\left(\dot{\bm{\zeta}}(t), \dot{\bm{\zeta}}(t)\right) \cdot k_{\text{sig}}\left(\dot{\bm{\zeta}}'(t), \dot{\bm{\zeta}}'(t)\right)}}\,.
    \label{eq:nomalised_signature}
\end{equation}
A higher similarity score indicates greater alignment in the kinematic profile and geometric structure of the adapted skill. We follow the implementation of \cite{barcelos2024path} and compute the signature with a truncation depth of 3, applied directly to the 3D linear and angular velocities separately. We use the default SE kernel, with a lengthscale chosen through a median heuristic. Furthermore, we use the left Jacobian of $SO(3)$ to provide the mapping from rotation vector's derivative to the angular velocities. 
The reward function combines task completion $r_{tc}$, signature similarity $r_{ss}$, and spatial and temporal penalties $r_{sp}$ and $r_{tp}$, respectively. $r_{tc}$ with a binary value contributes towards task success while $r_{ss}$ calculated with Eq.~\eqref{eq:nomalised_signature}, guides the learnt policy to retain high velocity similarities. The spatio-temporal penalties address the mismatch in skill execution time, specifically cases where the GP reaches the final pose too early, causing deviations from the desired goal. This encourages the policy to maintain the demonstration's velocity and acceleration profile.
\\
To define the penalties $r_{sp}$ and $r_{tp}$, let $\mathbf{p}_{\text{demo}}^n, \mathbf{q}_{\text{demo}}^n$ and $\mathbf{p}_{\pi}^n, \mathbf{q}_{\pi}^n$ denote the final position and orientation (in quaternions) of the demonstration and the policy-adapted GP trajectory, respectively. The spatial penalty is defined as
\begin{equation}
r_{sp} = \left\| \mathbf{p}_{\pi}^n - \mathbf{p}_{\text{demo}}^n \right\|_2 + 2 \cos^{-1} \left( \left| \left\langle \mathbf{q}_{\pi}^n, \mathbf{q}_{\text{demo}}^n \right\rangle \right| \right).
\label{eq:spatial_penalty}
\end{equation}
The temporal penalty is computed as the time difference between the final pose of the demonstration $\mathbf{p}_{\text{demo}}^{\text{n}}$ and the closest matching pose $\mathbf{p}_i$ in the policy-adapted trajectory. Given the demonstration length $T_{\text{demo}}$,
\begin{equation}
r_{tp} = T_{\text{demo}} - t^\pi; \quad \text{where } t^\pi = \arg\min_{t \in \mathcal{T}} \left\| \mathbf{p}_i - \mathbf{p}_{\text{demo}}^{\text{n}} \right\|_2.
\end{equation}
Thus, the full reward function is given by,
\begin{equation}
    r = r_{tc} + \alpha \cdot r_{ss} - \beta \cdot r_{sp} - \eta \cdot r_{tp},
    \label{eq:reward_eqation}
\end{equation}
where $\alpha, \beta$ and $\eta$ are scaling factors. We manually tune these values depending on the task to obtain the policy with the highest success rates and high velocity similarities.
\subsubsection{Frequent Checkpointing and Annealing}
With a skilled demonstration and a method to incorporate critical via-points, the RL agent's primary task is to adjust via-points to ensure task success whilst minimising velocity deviations. This biases the policy search toward regions near the conditioned GP for greater training stability. To enforce this constraint, frequent checkpointing is used to revert to the best-saved policy after a fixed number of episodes. Additionally, exploration annealing is applied by reducing the temperature over episodes, stabilising learning and preventing divergence.\begin{algorithm}[H]
\caption{Skill Adaptation with GPRL}
\label{algo:gprl_rl}
\begin{algorithmic}[1]
\Require Demonstration $\bm{\zeta}(t)$
\Ensure Adapted trajectory $\hat{\bm{\zeta}}(t)$
\State Initialise policy $\pi_\theta$, parameters $\theta$ and exploration schedule
\For{each episode $e$}
    \State Reset via-points to the demonstration set $\bm{\tilde{\Gamma}}$
    \State Condition the GP based on observed $\mathcal{TC}$
    \If{$N$th episodes after warm-up}
        \State Load best checkpoint: $\theta \leftarrow \theta^\star$
    \EndIf
    \State Propose corrective via-point deltas $\Delta\bm{\Gamma} \sim \pi_\theta$
    \State Update GP posterior using $ \bm{\Gamma^\ast} = {\bm{\tilde{\Gamma}}} + \Delta \bm{\Gamma}$
    \State Query adapted trajectory and kinematics
    \State Compute rewards $r_{tc}$, $r_{ss}$, $r_{sp}$ and $r_{tp}$
    \State Update policy parameters $\theta$
    \If{\textbf{current policy improves best performance}}
        \State Update best checkpoint: $\theta^\star \leftarrow \theta$
    \EndIf
\EndFor
\State Query the GP updated by the best policy to obtain $\hat{\bm{\zeta}}(t)$

\State \Return $\hat{\bm{\zeta}}(t)$
\end{algorithmic}
\end{algorithm}
\subsubsection{Training Details}
\begin{figure*}[!htbp]
    \centering
    \includegraphics[width=\linewidth, height=0.4\linewidth]{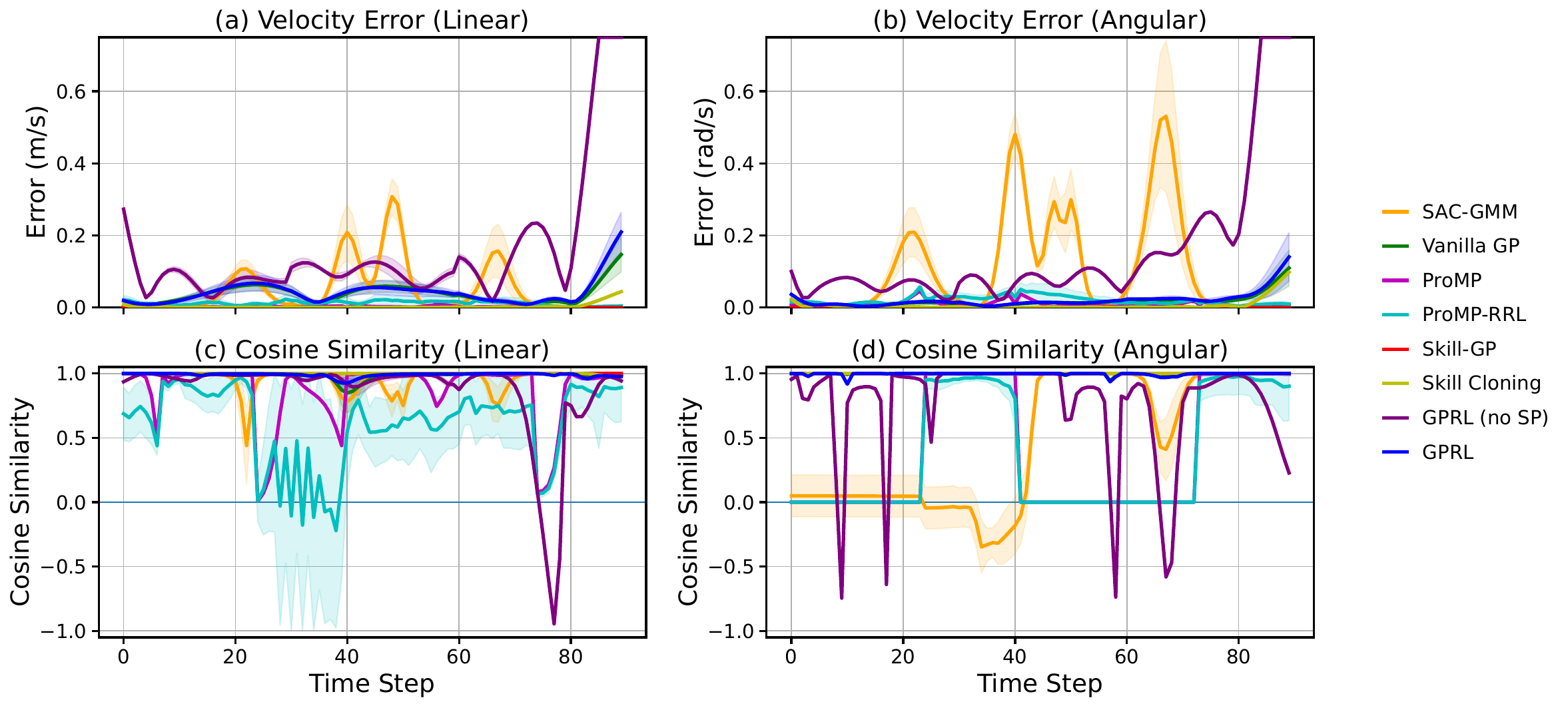}
    \caption{Retention of demonstrated velocities: Both Skill-GP and Skill Cloning demonstrate near-perfect retention. GPRL is able to obtain competitive scores in comparison to the vanilla GP. ProMP and ProMP-RRL show low magnitude error, but with poorer directional similarities.}
    \label{fig:similarity_res}
\end{figure*}
Given a single demonstrated trajectory $\bm{\zeta}(t)$ parameterised with our GP formulation, training proceeds episodically over varying task configurations.
At the beginning of each episode, the GP is conditioned on the current TC by aligning the closest via-point to the newly observed object location using Eq.~\eqref{eq:gpmean}, ensuring interpolation
through the observation. This conditioned GP provides a feasible initial trajectory, which the RL policy subsequently refines through corrective
shifts $\Delta \bm{\Gamma}$, of up-to $\pm4cm$, to the via-point set $\bm{\tilde\Gamma}$.
These via-points provide expressiveness of the trajectory and localised control over the trajectory shape. Too few via-points limit expressiveness, while too many increase the action space and hinder learning. In practice, for all the targeted tasks, 15 via-points per axis provide a good balance between representational fidelity and learning efficiency.
The corrective shifts update the GP model to obtain a new mean function $\bm{\hat{\mu}}(t)$. The updated GP is used to query the mean trajectory for execution, as well as velocities to calculate similarity rewards, before saving the transition in a replay buffer. At the end of each episode, the via-points are reset to their original state, $\bm{\tilde\Gamma}$, restoring the mean trajectory to $\bm{\mu}(t)$. After collecting a batch of transitions, the agent samples from the replay buffer to update the policy $\pi$ through Eq.~\eqref{eq:SAC_policy_update}. Through repeated updates over the collected batch, the policy gradually improves, maximising the cumulative rewards. Once the update is complete, a new round of interaction begins based on the updated policy. \rev{An algorithmic summary is provided in Algorithm~\ref{algo:gprl_rl}}.

\section{Experiments}
\begin{figure}
    \centering
    \includegraphics[width=1.0\linewidth, height=.25\linewidth]{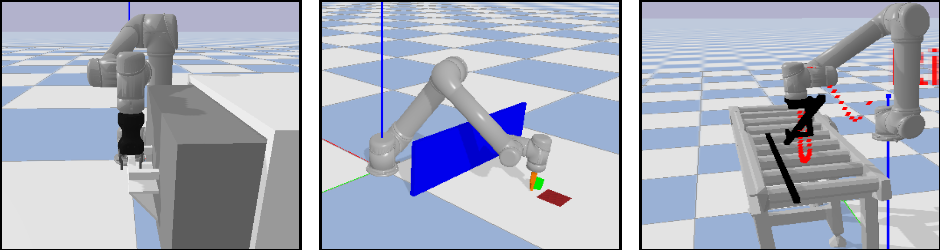}
    \caption{Simulated environments for drawer pulling, cube pushing and bar manipulation tasks.}
    \label{fig:task_setup}
\end{figure}

\subsection{Experimental Setup}
Our methods are first validated in simulation through three tasks \rev{(Fig.~\ref{fig:task_setup})} that assess their performance.
All experiments are conducted in PyBullet \cite{coumans2016pybullet}, with a single demonstration per task. The same setup is later replicated on hardware.
\\
\textbf{\textit{Drawer Opening (DOT)}}: The robot inserts its gripper between the cabinet door and handle, then pulls outward to open the drawer. The cabinet's position is varied across trials. 
\\
\textbf{\textit{Cube Pushing (CPT)}}: The robot pushes a cube across a flat surface while avoiding an obstacle and collisions. We test two variants: the static case, S-CPT, where the cube's initial pose is perturbed up to 20 cm along the xy-plane from the demonstration, and the dynamic case, D-CPT, where the cube's pose is continuously perturbed during execution. Success is defined by pushing the cube to the goal area.
\\
\textbf{\textit{Bar Manipulation (BMT)}}: The robot must grasp and remove one of two bars from a conveyor, compensating for up to 20 cm offset along the conveyor's width.

To transfer the learned policies to hardware, we leverage the proposed GP formulation to generate time-varying poses and analytical velocities for execution on the Universal Robots UR5e manipulator. For DOT, ArUco markers are used to localise the robot base, end-effector, task goal and drawer handle in the workspace, enabling the system to obtain their poses. A similar setup is followed for CPT. For BMT, we fine-tune a YOLOv11 \cite{Jocher_Ultralytics_YOLO_2023} model on 140 annotated images to detect and segment the bar. The colour images of the scene are obtained from a Intel Realsense D455 RGB-D camera, and the 3D points of the bar are recovered using depth data from the camera. Extrinsic calibration utilises an ArUco marker placed at a known transform from the base of the UR5e robot. 

\begin{figure}
    \centering
    \includegraphics[width=1.0\linewidth, height = 0.3\linewidth]{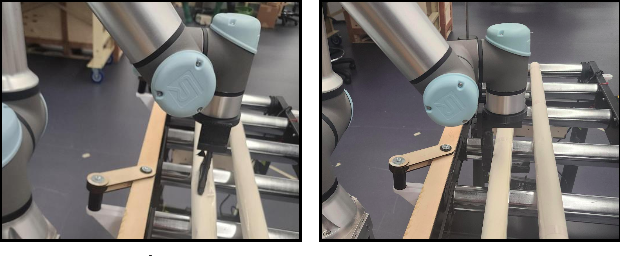}
    \caption{Vanilla GP crashes onto the bar (left), whereas our methods successfully adapt the learnt skill (right).}
    \label{fig:hw_res}
\end{figure}

\subsection{Baselines}
\rev{Our proposed methods are benchmarked against four state-of-the-art adaptation frameworks.}
\\
\textbf{\textit{SAC-GMM}}:
Following \cite{nematollahi2022robot}, SAC-GMM is implemented with checkpointing and annealing to aid in convergence.
\\
\textbf{\textit{Vanilla GP}}: The GP formulation used is described in Section~\ref{sec:proposed}, conditioned on the newly observed TC.
\\
\textbf{{\textit{Object-centric ProMP \cite{paraschos2013probabilistic}}}}: Demonstrations in the object frame are parameterised with a ProMP with 10 basis functions. At test time, we offset the observed TC from the nominal trajectory to obtain the adapted skill.
\\
\textbf{{\textit{ProMP-RRL \cite{carvalho2022adapting}}}}: We follow \cite{carvalho2022adapting} to first obtain the above object-centric ProMP and then use SAC to adapt the nominal skilled trajectory.

The default implementation of \cite{haarnoja2018soft} has been used for every benchmark that leverages SAC.

\subsection{Simulation Results}
Our simulated experiments comprise 100 runs with varying TCs, with deviations from the demonstration of up to 20 cm across all benchmarks. For SAC-GMM, ProMP and ProMP-RRL, we provide 20 demonstrations, linearly distributed over a 20 cm grid around the original demonstration to facilitate better representational power.

The results shown in Table \ref{tab:success_rates} show that SAC-GMM is unable to cope with the TC deviations despite access to multiple demonstrations.  
The vanilla GP, conditioned on the observed TC, achieves high success rates in simpler tasks such as DOT and CPT. However, its performance degrades in more complex tasks like BMT. 
A similar trend is observed for ProMP, which achieves a high success rate in all tasks except for in BMT. Conversely, ProMP-RRL does poorly in both S-CPT and D-CPT, where its policy is unable to avoid collisions with the plane but shows good performance in BMT.
In contrast, the success rate of our proposed methods, Skill-GP, Skill Cloning and GPRL, vastly outperforms the benchmarks for all tasks.

To evaluate the velocity retention for each method, we compute the cosine similarities of trajectory gradients and plot the velocity magnitudes relative to the demonstration. From Fig.~\ref{fig:similarity_res}, both Skill-GP and Skill Cloning produce near-perfect cosine similarities, with very low velocity errors. GPRL slightly underperforms compared to the vanilla GP in linear and angular velocity errors; it is still able to retain high cosine similarities demanded by more complex tasks and trajectories. In contrast, GPRL without similarity rewards (GPRL (no SP)) show the poorest similarity and the highest velocity errors, highlighting the importance of the similarity rewards.
The results also illustrate that ProMP and ProMP-RRL exhibit very low cosine similarity, despite achieving lower velocity errors than GPRL. Likewise, SAC-GMM performs poorly, with low cosine similarity and sporadically high velocity errors.

Lastly, we evaluate the inference time for each method. All methods were tested on an 11th-generation Intel Core i7-1185G7 processor with 16 GB DDR4 RAM. For BMT, on average, the Skill-GP converges at 1.8 seconds at each time-step. In contrast, both Skill Cloning and GPRL inference take only 0.03 seconds.

\begin{table}[htb]
    \centering
    \caption{Success Rates Across Experiments in Simulation}
    \label{tab:success_rates}
    \begin{tabular}{p{0.9cm}p{0.5cm}p{.5cm}p{0.5cm}p{0.5cm}p{0.7cm}p{0.7cm}p{0.7cm}}
\hline 
\rule{0pt}{2ex}    
 Task & SAC-GMM & Vanilla GP & ProMP & ProMP RRL & Skill-GP (Ours) & Skill Cloning (Ours) & GPRL (Ours) \\ \hline
 DOT & 28\% & 90\% & 92\% & 83\% & \textbf{100\%} & 95\% & 99\%
 \\ \hline
 S-CPT & 21\% & 83\% & 88\% & 41\% &\textbf{100\%} & 93\% & \textbf{100\%}
 \\ \hline
 D-CPT & 17\% & 71\% & 84\% & 16\% &\textbf{100\%} & 81\% & 95\% \\ \hline 
BMT & 0\% & 57\% & 39\% & 63\% &\textbf{100\%} & 85\% & 97\% \\ \hline
\end{tabular}
\end{table}
\subsection{Hardware Results}
Hardware results over 10 iterations \rev{(Table~\ref{tab:success_rates_hw})} show that SAC-GMM achieve poor success across all the tasks. ProMP performs better with high success rates except for in BMT. The addition of RL aids ProMP performance in BMT, but similar to simulation, the learnt policy is prone to collisions with the plane in both CPTs. Vanilla GP is able to adapt to the changing TCs to a degree, but fails in more complex tasks, especially when it is more than 10 cm away from the demonstration. In contrast, our methods achieve the highest success rates in all tasks. 
\rev{A video recording showing our experiments is available here: \url{https://youtu.be/CCgaULV3Uv0}}.
\begin{table}[htb]
    \centering
    \caption{Success Rates Across Experiments in Hardware}
    \label{tab:success_rates_hw}
    \begin{tabular}{p{0.9cm}p{0.5cm}p{.5cm}p{0.5cm}p{0.5cm}p{0.7cm}p{0.7cm}p{0.7cm}}
\hline 
\rule{0pt}{2ex}    
 Task & SAC-GMM & Vanilla GP & ProMP & ProMP-RRL & Skill-GP (Ours) & Skill Cloning (Ours) & GPRL (Ours) \\ \hline
 DOT  & 10\% & \textbf{100}\% & \textbf{100}\%& 70\% & \textbf{100}\% & \textbf{100}\% & \textbf{100}\%
 \\ \hline
 S-CPT & 20\% & 90\% & 80\%& 10\%& \textbf{100\%} & 90\% & \textbf{100\%}
 \\ \hline
 D-CPT & 10\% & 70\% & 70\% & 0\%&\textbf{100\%} & 90\% & \textbf{100\%} \\ \hline 
BMT & 0\% & 60\% & 20\%  & 50\% &\textbf{100\%} & 80\% & \textbf{100\%} \\ \hline
\end{tabular}
\end{table}
\subsection{Discussion}
\rev{Results show that GMM and ProMP-based adaptation frameworks underperform compared to GP-based methods, both in accuracy and kinematic retention. Despite access to multiple demonstrations, SAC-GMM lacks robustness to large TC deviations. Furthermore, GMM-based parameterisation offers no clear method to impose kinematic constraints for adaptation, relying instead on numerical approximations. This results in limited adaptability and large deviations in kinematic profiles. ProMPs offer a robust solution for skill adaptation, but their integration with RL is non-trivial. Although augmenting the nominal ProMP skill with residuals improves success rates in BMT, this compromises velocity similarity (Fig.~\ref{fig:similarity_res}). Additionally, residual learning causes jittery motions, often leading to collisions when operating close to surfaces.}
\begin{table}[b]
\centering
\caption{Effects of TC Offsets on Skill-GP Time}
\label{tab:convergence_times}
\begin{tabular}{c | c c c c c c}
\hline
Offset (cm) & $-15$ & $-10$ & $-2.5$ & $+2.5$ & $+10$ & $+15$ \\
Time (s)    & 2.96  & 2.28  & 1.18   & 0.93   & 1.97  & 2.45  \\
\hline
\end{tabular}
\end{table}
In contrast, vanilla GP conditioned on a newly observed TC can adapt effectively in simpler tasks. However, in more complex scenarios like BMT, where fine adjustments and minimisation of the velocity deviations are required for successful removal, it underperforms. A key limitation of the vanilla GP is that conditioning on distant TCs introduces spikes in the kinematic profile, which can cause failures \rev{(Fig.~\ref{fig:hw_res})}. 

Skill-GP achieves perfect success rates and near-perfect velocity retention across all tasks. Its only limitation is its convergence time,
which increases with the distance between the new TC and training demonstrations (see Table~\ref{tab:convergence_times}). This limits its applicability in real-world deployments. Despite this, Skill-GP offers a highly accurate solution in applications without real-time adaptation requirements. Skill Cloning eliminates the need for online optimisation by training a policy based on a dataset generated using Skill-GP. While this accelerates inference, performance depends on the dataset quality and scale, which is time-consuming to build. With 30000 expert state-action pairs, the cloned policy demonstrates near-perfect kinematic retention. However, without larger datasets or online refinement, it lacks robustness in terms of success rates. Thus, tasks that prioritise kinematic profile retention over accuracy with real-time adaptation can take advantage of this method.

These findings motivate GPRL, which introduces a balance between success rate and velocity retention. Unlike Skill Cloning, GPRL does not require a pre-existing dataset. Frequent checkpointing and annealing enable rapid convergence to a stable policy that achieves high success rates while retaining the required velocities. These trends are consistent with hardware results, where GPRL achieves high success rates for all tasks. 
Finally, we note some limitations. The contact tasks presented in this work involve relatively lightweight objects, and experiments with higher payloads have not yet been conducted. Furthermore, GPs have limited extrapolation capability, leading to velocity deviations toward the end of skill execution. While reward shaping partially mitigates this effect, residual deviations remain (see Fig.~\ref{fig:similarity_res}). In addition, our evaluations primarily consider translational deviations of up to 20 cm in the xy-plane. Deviations beyond this range result in significant deformation to the adapted skill's kinematic profile.
\section{Conclusion}
This paper presents a GP-based skill learning and adaptation framework. We demonstrate how we can attain structured skill adaptation for complex tasks while retaining the kinematic profile of the demonstration through optimisation and behaviour cloning. Furthermore, we developed a fast online variant that utilises signature similarity rewards for adapting to a wide range of TCs and reducing reliance on expert demonstrations. Our results boast near-perfect success rates with high similarity scores, vastly outperforming the benchmarks. It further proves that GP parameterisation aids adaptation with RL and provides an elegant way of velocity profile retention. Future work will consider an analytical approach instead of reward shaping to achieve the desired success and similarity scores, as well as accommodating diverse demonstrator preferences.

\bibliographystyle{ieeetr}
\bibliography{references}

\end{document}